\documentclass[11pt]{article}
\usepackage[margin=1in]{geometry}
\usepackage{graphicx,amsmath,amssymb,booktabs,hyperref,xcolor}
\usepackage[T1]{fontenc}
\usepackage{natbib}
\hypersetup{colorlinks=true,linkcolor=blue,citecolor=blue,urlcolor=blue}

\newcommand{\fpf}{\phi}
\newcommand{\Nstarts}{96}

\title{What a Cross-Model Fixed-Point Census Can and Cannot Arbitrate About Repetition}

\author{Nicol\'as Vera Z\'u\~niga\\
Independent Researcher, Chile\\
\texttt{nicovera@quetru.cl}
}
\date{}

\begin{document}
\maketitle

\begin{abstract}
Two accounts of neural text degeneration coexist. One locates the cause in the training data:
repetition in the corpus produces repetition in the output, established by training models on
repetition-sorted data. The other locates it in the trained network, in copying circuits and
repetition features. The accounts have never been arbitrated across a broad cohort of
\emph{pretrained} models, because the causal work necessarily trains its own. We report an
observational cross-model measurement in a different currency: the fixed-point structure of a
model's own short-window argmax map, censused from $\Nstarts$ random two-token starts over $17$
off-the-shelf models --- always unprompted, which a companion paper shows is a scope condition
rather than a detail, since nine tokens of conditioning move this readout across most of its range
--- with the four-way structural class stable across census seeds on $17$ of $17$. Three exhibits. At fixed corpus (The Pile) and fixed scale, the class is \emph{not determined}:
across two size-matched tiers, \texttt{pythia} is a funnel while RWKV, Mamba and a second
transformer family are not, and both families hold their class across roughly an order of magnitude
of scale. Six of seven models in that ladder reach the \emph{same} endpoint token, and the models
that concentrate on it most strongly are among those that never stay there --- so what varies is not
where trajectories go but whether the destination self-continues. A corpus manipulation available
off the shelf, the deduplicated Pythia suite, does not change the class. And the corpus-side
\emph{inflow} term proposed for this phenomenon does not select our endpoints once frequency is
controlled, on English and on three further languages. We are explicit about what this cannot do: it
is observational and cannot refute a training intervention. Funnels themselves are common in our
cohort --- eight of seventeen models across seven families and five corpora --- so the limit is not
that the phenomenon might be one model's peculiarity; it is that within the single corpus where
training data can be held fixed, only one of the available families funnels, so we cannot show from
that subset that the split is corpus-independent.
\end{abstract}

\section{Introduction}\label{sec:intro}

\subsection{Two accounts, and why they have not been arbitrated}

Neural language models fall into repetitive loops under maximum-a-posteriori decoding, and there are
two families of explanation for it.

\paragraph{The data side.} Repetition in the training text produces repetition in the output.
\citet{fu2021repetition} derive this from corpus bigram structure, attributing degeneration to
\emph{high-inflow} words --- those whose probability sum over all preceding words is large --- and
state the thesis plainly: the repetition problem is caused by the language itself. The empirical
champion is \citet{li2023repetition}, who train GPT-2 on repetition-sorted shards of five datasets,
report a strong correlation between repetition rates in training and generated text, and demonstrate
causation by dropping out attention to repetitive words during training. They further argue that the
high-inflow account, the likelihood-objective account and the self-reinforcement account all reduce
to one factor: penalizing repetitions in data.

\paragraph{The weights side.} Repetition is a property of the trained network. Copying circuits are
the named mechanism, and the two sides have been connected: \citet{hernandez2022repeated} show that
repeated training data disproportionately damages induction heads, and \citet{aoyama2026induction}
show that surface bigram repetition frequency governs whether induction heads form at all.

\paragraph{The gap is the cohort, not the connection.} We do not claim the accounts cannot see each
other --- they cite each other and cross repeatedly. What is absent is a broad cohort of
\emph{independently pretrained} models measured on a common readout. The causal work necessarily
trains its own models, which is what makes it causal and also what makes it few-model and
single-recipe. \citet{li2023repetition} say the remaining gap themselves: the model architecture and
size may also contribute, but the two factors have not been quantitatively evaluated.

\subsection{What this paper is, stated before the results}

This is \textbf{observational cross-model evidence in a currency different from a training
intervention}. It cannot refute \citet{li2023repetition}: they manipulated training data and
measured the consequence, and we did not. Our claim is narrower and of a different kind --- that on
a task-free structural readout, holding corpus and scale fixed does not fix the outcome.

We are equally explicit about the sharpest limit, and it goes here rather than in a limitations
paragraph --- but it is narrower than it first appears, and stating it loosely would misdescribe our
own data. Funnels are neither rare nor peculiar to one model: \textbf{eight of our seventeen models
are funnels, spanning seven families and five distinct training corpora}, with $\fpf$ from $0.328$ to
$1.000$.
What is scarce is not the phenomenon but a \emph{second funnel family within one corpus}. The only
corpus for which we can hold training data fixed across families is The Pile, and of the families
available there exactly one funnels. So the split reported in \S\ref{sec:e3} is not in doubt, and
neither is the existence of funnels across many unrelated models --- what we cannot establish from
that subset alone is that the split is \emph{corpus-independent}.

\section{Setup}\label{sec:setup}

\paragraph{The readout, and what is already published.} Iterate the model's own two-token
conditional deterministically, $x_{t+1} = \arg\max_x p(x \mid x_{t-1}, x_t)$, from $\Nstarts$ random
two-token starts, and census where trajectories land. The readout is the \emph{fixed-point fraction}
$\fpf$, the proportion of starts terminating in a token that maps to itself. \textbf{This map, and
the contrast between a model with a dominant attracting fixed point and one without, are established
in our own earlier work} \citep{veraz2026probes}, which identifies the attracting fixed point of the
argmax map and contrasts a model sending most starts to the newline token against one with no such
point. We claim neither here. What is new is the scale of the census, the class taxonomy and its
stability, and the corpus-versus-weights attribution the census makes possible.

\paragraph{Scope: a short-window readout, by nature and not by choice.}
A companion paper establishes that this readout \emph{disappears} as the window widens:
generalising the map so the state is the last $W$ tokens, raw $\fpf$ falls to $0.000$ on four of six
models by $W = 16$ \citep{veraz2026domain}. Everything below is therefore a statement about how a
model reads a \emph{fragment}. We put this in Setup rather than in Limits because it is a property of
the construction rather than a shortfall of the evidence, and because a reader who takes our
$\fpf$ for a general property of the model would be taking it for something it is not.

\paragraph{Scope: the raw domain, and only it.}
Every census reported here conditions on nothing: the two-token starts are drawn at random and no
prefix precedes them. That is a second scope condition, and the same companion paper is what makes
it load-bearing rather than a technicality --- nine tokens of conditioning move $\fpf$ across most of
its range and change the four-way class, while an instruction-tuning intervention worth $60.5$
IFEval points moves the class by zero \citep{veraz2026domain}. The classes below are therefore
properties of \emph{a model under this condition}, not of the model simpliciter, and a reader who
carries one of our class labels to a prompted setting is carrying it somewhere it has been measured
not to hold. We state both scopes here, together, because they bound every number in the paper.

\paragraph{Classes.} Trajectories are classified \textsc{funnel} (one dominant attracting fixed
point), \textsc{none} (cycles or wandering), \textsc{fragmented} (many small basins) and
\textsc{borderline}. The \emph{modal share} of a census is the fraction of the $\Nstarts$ starts whose
trajectory ends at the single most common endpoint token; it measures how concentrated the
destinations are, independently of whether any of them is a fixed point. Class is a deterministic
function of the pair $(\fpf, \text{modal share})$ under a rule fixed before the data was seen:

\begin{center}\small
\begin{tabular}{ll}
\toprule
class & rule \\
\midrule
\textsc{funnel}     & $\fpf \geq 0.30$ and modal share $\geq 0.30$ \\
\textsc{none}       & $\fpf \leq 0.10$ \\
\textsc{fragmented} & $\fpf \geq 0.30$ and modal share $< 0.20$ \\
\textsc{borderline} & anything else, reported and never forced into a class \\
\bottomrule
\end{tabular}
\end{center}

\noindent The gap matters for \S\ref{sec:e3}, so we state it rather than leave it to be
reconstructed: \textsc{funnel} and \textsc{none} are separated by an entire unoccupied band, and the
models the paper contrasts sit far from both edges of it. No boundary in this rule falls between the
values being compared. The funnel geometry --- many states
feeding one self-continuing token --- was derived theoretically by \citet{fu2021repetition}, whose
inflow analysis explains why trajectories concentrate; the classes here are that geometry measured
on a model's own conditional rather than on corpus counts.

\paragraph{Stability, which licenses everything else.}
Across $17$ off-the-shelf pretrained models censused at two independent seeds, the class is stable on
$17$ of $17$. The modal endpoint token is stable on $15$ of $17$; the two exceptions are reported
and excluded from every comparison rather than assigned a class.

\paragraph{Why this readout and not a generation statistic.} Nothing is sampled and nothing is
generated. There is no decoding temperature, no prompt, and no text to score. That makes the
measurement cheap enough to run over many models, and it makes it a property of the conditional
rather than of a generation procedure --- but it also means our quantity is not the $\mathrm{rep}\text{-}n$
of the literature, and no result here transfers to that quantity without an argument we do not make.

\section{Related work: an adjacent measurement, and what separates it}\label{sec:related}

The dynamical-systems framing of degeneration is not ours and we do not claim it.
\citet{du2025correlation} describe degeneration as a collapse from a higher-dimensional trajectory
onto a lower-dimensional attractor, borrowing explicitly from chaos theory, and measure a
degeneration-detecting dynamical property --- the correlation dimension of next-token
log-probability trajectories --- across several independently trained families. That is the closest
published object to the present one, and the differences are worth stating precisely rather than
leaving implicit. Their state space is continuous log-probability vectors seeded from real text;
ours is a deterministic map over discrete token space seeded from random pairs. Their quantity is a
dimension estimated over a trajectory; ours is the fraction of starts reaching a token that maps to
itself. Neither measurement is a special case of the other, and the vocabulary they use predates and
is independent of our use of it.

Same-corpus, size-matched comparisons of pretrained models across architecture families are likewise
an established design rather than a new one, and we adopt it: \citet{michaelov2024recurrent} compare
Pythia, RWKV and Mamba checkpoints trained on The Pile at matched weight classes, and
\citet{wang2025universality} compare Pythia against Mamba at matched size on induction circuits. Our
contribution on that axis is a readout, not a design. Both report substantial cross-architecture
\emph{similarity}, and \citet{michaelov2025phases} report consistency of behavioural phases across
architecture, training data and scale. \S\ref{sec:e3} is therefore a counterexample on a readout
those studies do not use, and is presented as one.

\section{E3: at fixed corpus and fixed scale, the class is not determined}\label{sec:e3}

\begin{table}[t]\centering
\small
\begin{tabular}{llrrrl}
\toprule
tier & model & family & $\fpf$ & modal share & modal endpoint \\
\midrule
400M & \texttt{pythia-410m}      & GPTNeoX & \textbf{0.458} & 0.453 & \verb|\n| \\
400M & \texttt{rwkv-4-430m-pile} & RWKV    & 0.010 & 0.521 & \verb| time| \\
400M & \texttt{mamba-370m-hf}    & Mamba   & 0.010 & 0.823 & \verb| first| \\
\midrule
150M & \texttt{pythia-160m}      & GPTNeoX & \textbf{0.432} & 0.432 & \verb|\n| \\
150M & \texttt{gpt-neo-125m}     & GPTNeo  & 0.052 & 0.464 & \verb| side| \\
150M & \texttt{rwkv-4-169m-pile} & RWKV    & 0.000 & 0.474 & \verb|\n| \\
150M & \texttt{mamba-130m-hf}    & Mamba   & 0.000 & 0.573 & \verb|The| \\
\bottomrule
\end{tabular}
\caption{Two size-matched tiers, every model trained on The Pile. Within a tier the largest and
smallest parameter counts are within $1.16\times$ (400M) and $1.35\times$ (150M) of each other, a
bound fixed before the runs. Both census seeds agree on the class for all seven. Source: F178,
\texttt{results/size\_matched\_pile.json}.}
\label{tab:tiers}
\end{table}

Table~\ref{tab:tiers} holds the corpus fixed and the weight class fixed, and the class still varies.
\texttt{pythia} is a funnel in both tiers; RWKV, Mamba and \texttt{gpt-neo} are not.

\paragraph{It is not a scale effect, and that was the pre-registered risk.}
An earlier version of this comparison used \texttt{gpt-neo-2.7B} against \texttt{pythia-410m}, which
confounds family with a $6.6\times$ size gap. The decisive cell was registered in advance:
\texttt{gpt-neo-125m} against \texttt{pythia-160m}, $1.28\times$ apart on the same corpus, with the
condition that if they landed in the same class the claim would be withdrawn. They do not ---
$\fpf = 0.052$ against $0.432$, both stable across seeds.

\paragraph{We do not say ``architecture'', and the reason is in the table.}
\texttt{gpt-neo} is a transformer, from the same laboratory, trained on the same corpus, and it sits
with the recurrent models rather than with \texttt{pythia}. The split is therefore not
transformer-versus-recurrent, and naming it after architecture would be reading the table wrongly.
What the evidence supports is the weaker and more precise statement: \textbf{at fixed corpus and
fixed scale the class is not determined}, and something on the weights side that we have not
identified decides it.

\paragraph{The domain is held fixed here too, and saying so is not a formality.}
Every cell in Table~\ref{tab:tiers} is a raw-domain census, per Setup. That third fixed condition
deserves naming beside the other two, because unlike them it is known to be capable of moving the
quantity: nine tokens of conditioning change this class in the companion measurement
\citep{veraz2026domain}. So what E3 establishes is non-determination \emph{at one domain}. Whether
these same seven models separate the same way under a prefix is a different experiment, not a
corollary of this one, and we do not run it.

\begin{table}[t]\centering
\small
\begin{tabular}{lrlrrl}
\toprule
family & size & class & $\fpf$ & modal share & modal endpoint \\
\midrule
Pythia  & 70M    & \textsc{funnel} & 0.802 & 0.792 & \verb|\n| \\
Pythia  & 160M   & \textsc{funnel} & 0.432 & 0.432 & \verb|\n| \\
Pythia  & 410M   & \textsc{funnel} & 0.458 & 0.453 & \verb|\n| \\
Pythia  & 1000M  & \textsc{funnel} & 0.365 & 0.406 & \verb|\n| \\
\midrule
GPT-Neo & 125M   & \textsc{none}   & 0.052 & 0.464 & \verb| side| \\
GPT-Neo & 1300M  & \textsc{none}   & 0.000 & 0.536 & \verb|\n| \\
GPT-Neo & 2700M  & \textsc{none}   & 0.036 & 0.594 & \verb|\n| \\
\bottomrule
\end{tabular}
\caption{Both families hold their class across scale on one corpus: Pythia over a $14\times$ span,
GPT-Neo over $22\times$. No model is class-unstable across seeds. Source: F179,
\texttt{results/family\_scale\_ladder.json}.}
\label{tab:ladder}
\end{table}

\paragraph{The split is a stable family property, not two unrepresentative checkpoints.}
Table~\ref{tab:ladder} extends each family across scale. Neither ladder breaks.

\paragraph{The sharpest observation is in the endpoint column.}
Six of the seven models in Table~\ref{tab:ladder} reach the \emph{same} endpoint token, the newline
--- every Pythia, and GPT-Neo at $1300$M and $2700$M. GPT-Neo concentrates on it \emph{more} strongly
than some Pythias do: modal share $0.536$ and $0.594$ against \texttt{pythia-160m}'s $0.432$. Yet
GPT-Neo's $\fpf$ is $0.000$ and $0.036$ while every Pythia exceeds $0.36$. The same holds inside
Table~\ref{tab:tiers} at matched size: \texttt{rwkv-4-169m-pile} and \texttt{pythia-160m} share the
newline endpoint at nearly the same concentration, $0.474$ against $0.432$, with $\fpf$ of $0.000$
against $0.432$.

\textbf{Where trajectories go is shared; whether the destination self-continues is not.} The models
that funnel hardest onto the newline are among those that never stay on it.

\paragraph{The decomposition is arithmetic, not a reading of the table.}
Compare $\fpf$ against the modal share row by row in Table~\ref{tab:ladder}. For every Pythia the two
agree to within four census starts out of $\Nstarts$ --- $0.802$ against $0.792$, $0.432$ against
$0.432$, $0.458$ against $0.453$, $0.365$ against $0.406$. For every GPT-Neo the share exceeds $\fpf$
by forty to fifty-four starts. In other words $\fpf$ factorises, to within the census resolution, as
\[
  \fpf \;\approx\; (\text{mass arriving at the modal endpoint}) \times (\text{whether that endpoint self-continues}),
\]
with the second factor near $1$ for one family and near $0$ for the other while the first is
comparable across both. The corpus appears to fix the first term and something on the weights side
the second. We state this as a factorisation others can test rather than as an impression, and
\S\ref{sec:conclusion} asks for exactly that test.

\section{E1: an off-the-shelf corpus manipulation that does not change the class}\label{sec:e1}

The Pythia suite was released in duplicate, one copy trained on The Pile and one on a
near-deduplicated copy, explicitly so that deduplication could be studied
\citep{biderman2023pythia}. That is the nearest thing to the data-side manipulation available without
training anything.

Both members remain \textsc{funnel}: $\fpf = 0.458$ for \texttt{pythia-410m} and $0.427$ for
\texttt{pythia-410m-deduped}, with the same modal endpoint token.

\paragraph{This pair does not differ only in deduplication, and the caveat is not incidental.}
The deduplicated Pile is approximately $207$B tokens while both suites are trained to approximately
$300$B \citep{biderman2023pythia}, so the deduplicated models run about $1.45$ epochs and re-encounter
a large fraction of their corpus a second time. \textbf{In a paper about repetition this matters}: the
manipulation re-introduces the very variable the data-side account identifies as causal. We therefore
do not present this as a clean corpus manipulation. It is a null on a readout with no prior
expectation of invariance, from a pair whose difference is smaller than its label suggests.

\paragraph{And a null here is weaker than it looks for a second reason.}
\citet{hernandez2022repeated} show that damage from repeated data is strongly non-monotonic --- a
double-descent, concentrated in a particular range of repetition frequency. If the Pile's duplication
does not fall in that range, the data-side account predicts no effect either, and our null is
consistent with both accounts rather than discriminating between them. We report E1 because it is the
manipulation the field would ask for, and we report what it does and does not establish.

\section{E2: the corpus-side inflow term does not select the endpoints, and this agrees with the data side}\label{sec:e2}

\citet{fu2021repetition} attribute degeneration to high-inflow words, where the inflow of a token is
the probability sum over all its predecessors. If the endpoints our maps reach are the tokens that
account predicts, they should be high-inflow ones.

Thirteen of the seventeen models are readable for this comparison, and the four exclusions are of
three different kinds. Two (\texttt{LFM2-2.6B}, \texttt{starcoder2-3b}) disagree with themselves
across census seeds about which endpoint token is modal, so the quantity being predicted is not
stable and they are dropped before anything is computed. One (\texttt{bloom-3b}) has an endpoint
token that occurs \emph{zero} times in an English corpus --- a statement about the corpus, not about
the theory, and \S\ref{sec:e2} returns to it. One (\texttt{polyglot-ko-1.3b}) had a tokenizer that
would not load, which is an infrastructure failure and evidence about nothing. We keep the three
reasons apart because they license different inferences.

On an uncontrolled reading the endpoints are high-inflow: $12$ of the $13$ have an endpoint at or
above the $90$th inflow percentile, median $99.87$. That reading is an artefact of frequency.
\citet{fu2021repetition} themselves distinguish the two, arguing that it is not the high-frequency
words but the high-inflow words that lead to repetition, so we compare each endpoint against the $50$
tokens nearest it in log-frequency. Against that control the endpoint beats its matched peers in
$1$ of $13$ models, with a median matched percentile of $32.0$ against $50$ by construction.

The result is not an artefact of measuring every model on English. Three models whose endpoints are
not English were measured on their own languages at matched corpus size: $6.0$ on Spanish, $16.0$ on
Korean and $96.0$ on Japanese. Two of the three were unmeasurable in English --- their endpoint tokens
occur zero times in an English corpus --- and both land far below the null on the right corpus.

\paragraph{This agrees with the data-side account rather than opposing it.}
\citet{li2023repetition} already subsume the inflow account into repetition-in-training-data by a
controlled experiment: merging only the repetitive high-inflow pairs, a small fraction of training
words, matches the full high-inflow method, while merging random high-inflow pairs of the same size
does not. Inflow was demoted by the data side itself. Likewise \citet{michaelov2025phases} find that
unigram frequency dominates a large share of word-level behavioural variance across pretrained
models. \textbf{E2 is therefore a consistency check, not a refutation}: our maps funnel to
\emph{common} tokens rather than to high-inflow ones, which is what both of those results predict.

\section{Limits}\label{sec:limits}

\paragraph{The currency gap, first because it is the largest.} \citet{li2023repetition} manipulated
training data and measured the consequence. We measured pretrained models and compared them. An
observational contrast across models cannot refute a controlled intervention, and nothing here should
be read as doing so. Where our results bear on theirs at all, as in \S\ref{sec:e2}, they agree.

\paragraph{One funnel family \emph{within the fixed-corpus subset}.} A reader may worry that
\S\ref{sec:e3} rests on a peculiarity of one model suite. Our own cohort answers that: funnels appear
in seven families across five corpora (\S\ref{sec:intro}), so the phenomenon is common and is not a
Pythia artefact. The genuine limit is more specific. Within The Pile --- the one corpus where we can
vary the family with training data held fixed --- only Pythia funnels among the families available to
us. A second funnel family at fixed corpus would let us show the split is corpus-independent rather
than a fact about which families happen to be Pile-trained; the obvious public candidate exceeded the
memory available to us, and we did not obtain one. \textbf{That is what \S\ref{sec:e3} cannot
currently show, and it is narrower than "the effect may not be real".}

\paragraph{What ``not determined'' does and does not mean.} \S\ref{sec:e3} holds corpus and parameter
count fixed. It does not hold fixed the training schedule, the optimiser, the data order, the
tokenizer, or the many choices that differ between independently trained families. A difference under
those conditions means \emph{not explained by corpus or scale}. It does not identify what does explain
it, and we make no attempt to.

\paragraph{Readout scope.} $\fpf$ is a property of a deterministic two-token map. It is not
$\mathrm{rep}\text{-}n$ on generated text, and the relationship between them is unestablished. A
model with no fixed points in this census may still degenerate under sampling, and we have not
checked.

\paragraph{Statistical refusals, recorded before the numbers.} No $p$-values are reported. The cohort
is $17$ models and the exhibits rest on two families; a significance test at that $n$ could not fail
informatively. No rank correlations are computed between class and scale. Where a comparison could not
be made cleanly --- two models whose modal endpoint is unstable across seeds, models whose endpoints
do not occur in a given corpus --- the models are named and excluded rather than assigned a value.

\section{Conclusion}\label{sec:conclusion}

On a deterministic, task-free readout of a model's own conditional, holding the training corpus fixed
and the parameter count fixed does not fix the fixed-point structure. Two families trained on the
same corpus keep opposite classes across an order of magnitude of scale, and a corpus manipulation
released expressly to study deduplication does not move the class either --- though that pair is
confounded by a second pass over much of its data, and we say so. Meanwhile the corpus-side term
proposed for this phenomenon does not select the endpoints once frequency is controlled, in four
languages, which is what the data-side literature now predicts for itself.

The observation we would most like tested by others is the decomposition in \S\ref{sec:e3}: across
families, scales and corpora, models agree remarkably about \emph{which} token trajectories arrive at
and disagree about whether that token maps to itself. Six of seven models in one ladder land on the
newline; the ones that concentrate on it most strongly are the ones that never stay. If that survives
a cohort with more than one funnel family in it, it is a fact about language models that neither
account currently explains.

\bibliographystyle{plainnat}
\bibliography{refs}

\end{document}